\documentclass[sn-mathphys,Numbered]{sn-jnl}

\usepackage{graphicx}%
\usepackage{multirow}%
\usepackage{amsmath,amssymb,amsfonts}%
\usepackage{amsthm}%
\usepackage{mathrsfs}%
\usepackage[title]{appendix}%
\usepackage{xcolor}%
\usepackage{textcomp}%
\usepackage{manyfoot}%
\usepackage{booktabs}%
\usepackage{algorithm}%
\usepackage{algorithmicx}%
\usepackage{algpseudocode}%
\usepackage{listings}%

\usepackage{tabulary,times,caption,fancyhdr,amsbsy,latexsym}
\usepackage[utf8]{inputenc}
\usepackage{url,morefloats,floatflt,cancel,tfrupee,colortbl,pifont}

\begin{document}

\title[Article Title]{Dual skip connections in U-Net, ResUnet and U-Net3+ for remote extraction of buildings}


\author*[1,2]{\fnm{Bipul} \sur{Neupane}}\email{bneupane@student.unimelb.edu.au}

\author[1,2]{\fnm{Jagannath} \sur{Aryal}}\email{jagannath.aryal@unimelb.edu.au}

\author[2]{\fnm{Abbas} \sur{Rajabifard}}\email{abbas.r@unimelb.edu.au}

\affil[1]{\orgdiv{Earth Observation and AI Research Group, Faculty of Engineering and IT}, \orgname{The University of Melbourne}, \orgaddress{\city{Melbourne}, \postcode{3053}, \state{VIC}, \country{Australia}}}

\affil[2]{\orgdiv{Department of Infrastructure Engineering, Faculty of Engineering and IT}, \orgname{The University of Melbourne}, \orgaddress{\city{Melbourne}, \postcode{3053}, \state{VIC}, \country{Australia}}}


\abstract{
Urban buildings are extracted from high-resolution Earth observation (EO) images using semantic segmentation networks like U-Net and its successors. Each re-iteration aims to improve performance by employing a denser skip connection mechanism that harnesses multi-scale features for accurate object mapping. However, denser connections increase network parameters and do not necessarily contribute to precise segmentation. In this paper, we develop three dual skip connection mechanisms for three networks (U-Net, ResUnet, and U-Net3+) to selectively deepen the essential feature maps for improved performance. The three mechanisms are evaluated on feature maps of different scales, producing nine new network configurations. They are evaluated against their original vanilla configurations on four building footprint datasets of different spatial resolutions, including a multi-resolution (0.3+0.6+1.2m) dataset that we develop for complex urban environments. The evaluation revealed that densifying the large- and small-scale features in U-Net and U-Net3+ produce up to 0.905 F1, more than TransUnet (0.903) and Swin-Unet (0.882) in our new dataset with up to 19x fewer parameters. The results conclude that selectively densifying feature maps and skip connections enhances network performance without a substantial increase in parameters. The findings and the new dataset will contribute to the computer vision domain and urban planning decision processes.
}

\keywords{building extraction, convolutional neural networks, computer vision, u-net, remote sensing, semantic segmentation}



\maketitle

\section{Introduction}\label{sec1}
Semantic segmentation of EO images is the current state-of-the-art (SOTA) method for urban feature extraction and land use and land cover classification (LULC) \cite{neupane2021deep}. The evolving nature of semantic segmentation has seen traditional classifiers including edge-based, shadow-based, region-based, and machine learning approaches \cite{hossain2019segmentation}. These methods lack precision due to their dependency on mid-level semantic characteristics (hand-crafted features). Neural networks overcome this limitation with an increased number of ``layers'' and provide contrasting results when compared to machine learning predecessors like random forest \cite{du2015semantic} and support vector machine \cite{huang2012svm}. An encoder-decoder network (EDN) with a convolutional neural network (CNN) as its fundamental backbone (encoder) has become the common network configuration for semantic segmentation \cite{minaee2021image}. U-Net \cite{ronneberger2015u} is the most common EDN for semantic segmentation with its legacy in all application domains ranging from the segmentation of medical imagery to satellite imagery. It is also the most revisited and modified network for performance gain. In this paper, we rethink the architecture of three widely appreciated EDNs $-$ U-Net, ResUnet, and U-Net3+ $-$ particularly the ``skip connection'' components of the networks that transport the context information learned by CNN encoder to the higher-resolution layers of the decoder for precise localisation. 


Skip connections are pivotal components of EDNs like U-Net and its re-iterations of evolving nature. The connections serve as a bridge to pass the low-level information from the encoder of U-Net to its decoder to make the utmost use of the encoder-decoder configuration. The encoder is essentially a CNN that generates multi-scale feature maps of the input image using convolutions and down-sampling operations. The decoder is symmetrical to the encoder with up-sampling operations that replace the down-sampling operations, comprising high-level and coarse-grained semantic information. With skip connections, U-Net uses the vectors of the feature maps generated in the encoder to map the objects into the final segmented maps, providing an effective trade-off between the use of context and precise localisation. ResUnet \cite{zhang2018road} replaces the convolutional layers in U-Net with residual blocks to allow effective training with skip connections (aka. residual connections) between multi-scale feature maps within its encoder. These connections pass not only the feature maps but also the gradient information, which is beneficial when dealing with deeper networks. U-Net++ \cite{zhou2019unet++} uses nested and dense skip connections to reduce inter-encoder-decoder semantic gaps for increased precision in segmentation. The dense connections in ResUnet and U-Net++ significantly increase the network parameters. U-Net3+ \cite{huang2020unet} reduces the network parameters with an efficient way of leveraging the multi-scale features with full-scale skip connections. The inter-encoder-decoder skip connections of U-Net are re-designed, and new intra-connections between the decoder layers are added to capture fine-grained details and coarse-grained semantics in full scales. The full-scale skip connection assumes equal weights among the feature maps generated at different scales. However, different scales of feature maps possess different levels of discrimination and context information, which is not realised in the U-Net3+ configuration. To sum up, the newer iterations of U-Net use denser skip connections for performance gain with more effective solutions. However, recent studies \cite{wang2022uctransnet} have found that the skip connections do not always provide performance gain, and our finding suggests that densifying the selective feature maps and skip connections can provide performance gains without a significant increase in network parameters. 

To address the aforementioned weaknesses of the U-Net and its re-iterations, we propose a dual skip connection mechanism (DSCM) for U-Net, a dual respath skip connection mechanism (DRSCM) for ResUnet, and a dual full-scale skip connection mechanism (DFSCM) for U-Net3+. Unlike the existing similar mechanisms in DR-Unet \cite{le2021dr} and DPN-Unet-TypeII \cite{xu2021dual}, our mechanisms deepen the selective feature maps of the networks and double the information passing through the skip connections. DSCM in U-Net deepens the convolutional blocks of different scales and DRSCM on ResUnet deepens its residual block and doubles the residual skip connections. In U-Net3+, we introduce a new multi-scale feature aggregation technique to integrate the DFSCM. Unlike, the aggregation that assumes equal weights for all feature maps in U-Net3+, the proposed DFSCM gives higher weight to the small-scale feature maps, where the context information is still intact. A lower weight is given to the large-scale features as they lose context with downsampling operations. Such denser skip connections have not been studied in U-Net3+ to the best of our knowledge. To study the effects of dual skip connections in different scale layers, we propose and experiment with three scale variants of each network. With the three variants of the three connection mechanisms on the three networks, a total of nine new network configurations are proposed and studied in this paper.

The experimental design of the proposed networks is categorised in terms of the spatial resolution of datasets. Four datasets of VHR and high-resolution types are used for the evaluation, with one newly developed multi-resolution dataset of complex urban building samples. The performance of the nine new networks is compared to the vanilla versions of U-Net, ResUnet, and U-Net3+ and other similar SOTA networks like U-Net++ \cite{zhou2019unet++}, TransUnet \cite{chen2021transunet}, Swin-Unet \cite{cao2022swin}, Attention U-Net \cite{oktay2018attention}, U$^2$Net \cite{qin2020u2}, FCN8s \cite{long2015fully}, DeepLabv3+ \cite{chen2017deeplab}, and SegNet \cite{badrinarayanan2017segnet}. All of the proposed and vanilla networks are evaluated with four evaluation measures. The contributions of this paper are:

\begin{enumerate}
    \setcounter{enumi}{0}
    \item Proposition of DSCM and DRSCM for U-Net and ResUnet to deepen the encoder and allow denser skip connections for an effective trade-off between the use of context and precise localisation of building footprints from VHR and high-resolution images.
    \item Proposition of DFSCM for U-Net3+ with a new multi-scale feature aggregation technique to concatenate feature maps of different scales with different weights. Increased weight is given to the smaller-scale feature maps, where the context information is intact.
    \item Comprehensive comparison of the proposed mechanisms on different scale layers of U-Net, ResUnet, and U-Net3+ with a total of nine new network configurations. The nine networks are tested on four datasets of different spatial resolutions. An ablation study is provided to find the scale layers in U-Net that provide the highest performance gain with the proposed DSCM.
    \item Development of a new multi-resolution dataset for robustness check with urban building samples of different spatial resolutions. The dataset is developed in an end-to-end manner using APIs to collect imagery and create labels, without manual annotation.
\end{enumerate}

\section{Related Work} \label{sec:relw}

\subsection{CNNs, FCNs and EDNs} \label{sec:relw:CNNs}
The earliest forms of CNN are AlexNet \cite{krizhevsky2012imagenet}, VGGNet \cite{simonyan2014very}, GoogleNet \cite{szegedy2015going}, ResNet \cite{he2016deep}, and Xception \cite{chollet2017xception}. They are now used to extract multi-scale feature information from images \cite{chen2018aerial,ayala2021deep} for segmentation purposes in FCNs and EDNs like SegNet \cite{badrinarayanan2017segnet} and U-Net \cite{ronneberger2015u}. Some of the earliest studies of CNN-based building footprint extraction \cite{mnih2013machine,saito2016multiple} perform patch-based segmentation supported by post-processing methods to improve accuracy and precision. CNN-based pixel-level building segmentation is achieved by Mask R-CNN that uses CNNs like ResNet as feature extractor (aka. backbone) \cite{zhao2018building,griffiths2019improving}. Similarly, building footprint extraction has also seen the use of FCN \cite{maggiori2016convolutional} and different versions of it: FCN2s, FCN4s, and FCN8s \cite{zhong2016fully,yang2018building}.

The early forms of EDNs (SegNet and U-Net) were first introduced to segment indoor/outdoor scenes and medical images. They followed their way to EO images including multi-spectral images from WorldView-3 \cite{li2019semantic}, synthetic aperture radar (SAR) and multi-spectral images from Sentinel-1 and 2 \cite{ayala2021deep} for building segmentation. Some have also achieved building classification \cite{pan2020deep}. SegNet is also experimented with further modifications \cite{bischke2019multi,sariturk2020feature} and in combination with U-Net to form Seg-Unet \cite{abdollahi2022ensemble}. Similarly, ResUnet \cite{zhang2018road} is introduced to extract road networks with residual units added to U-Net to ease the training and further enrich the skip connections. ResUnet is further experimented for building segmentation with multi-modal hand-crafted features \cite{xu2018building} and ultra-high resolution (UHR) images \cite{li2019semantic}. Due to its legacy in the domain of computer vision, widely used versions of the U-Net family are the focus of this paper.

\subsection{The legacy and evolution of U-Net} \label{sec:relw:unet}
The popularity of U-Net \cite{ronneberger2015u} comes from addressing the loss of context information in the CNNs due to (i) the heavy use of pooling and max-pooling operations, and (ii) the imbalanced trade-off between localisation and the use of context. U-Net has seen an explosion in usage in medical imaging since its introduction and several variants have been proposed in this domain \cite{siddique2021u}. However, the loss of context information cannot be avoided while using CNNs as a feature extractor. U-Net is followed by a number of improvements that are focused on utilising the multi-scale feature maps of CNNs in an efficient manner. Among many, some of the successful variants are 3D U-Net \cite{cciccek20163d}, V-Net \cite{milletari2016v}, Attention U-Net \cite{oktay2018attention}, U-Net++ \cite{zhou2019unet++}, R2U-Net \cite{alom2018recurrent}, Inception-U-Net \cite{zhang2020dense}, ResUnet \cite{zhang2018road}, Dense U-Net \cite{wang2019dense}, adversarial U-Net\cite{schonfeld2020u}, U$^2$Net \cite{qin2020u2}, and U-Net3+ \cite{huang2020unet}. These U-Nets are widely used to segment both medical and EO images. The recent advancement in U-Net seems also to be affected by the rise of Vision Transformer (ViT) \cite{dosovitskiy2020image}, which brings the science of Transformer \cite{vaswani2017attention} from natural language processing to a computer vision problem. The concept of ViT tries to replace CNNs in semantic segmentation but has still not been able to fully replace it. The adaptation of ViT in the recent U-Net versions includes TransUnet \cite{chen2021transunet}, TransFuse \cite{zhang2021transfuse}, and Swin-Unet \cite{cao2021swin}. TransUnet builds upon the problem of CNNs failing to fully learn the global and remote semantic information interaction because of the involvement of the convolutional process. The prior concepts of feature pyramid \cite{lin2017feature}, DeepLab \cite{chen2017deeplab}, atrous convolution layers \cite{chen2018encoder}, context encoder network (CE-Net) \cite{gu2019net}, self-attention \cite{wang2018non}, and attention gate \cite{schlemper2019attention} also have tried to address this problem. With U-Net3+ as the most recent version based on the naming, there are some recent studies that have tried to improve the network with an addition/integration of attention module \cite{li2020macu}, residual unit \cite{qin2022improved}, and transformer \cite{chen2023improved}. O-Net \cite{wang2022net} is a recent development that realises the addition of a self-attention mechanism on the transformer and combining it with CNN can marginally improve the network when compared to computational-heavy networks based on ViT. The improvement in performance is vitally important to computer vision but raises some serious questions regarding the computation expense. Unlike computationally super-expensive transformer-based semantic segmentation to address the problem of inadequate learning of global and remote semantic information, some other advancements in U-Net variants seek to lower the number of network parameters in U-Net while improving the performance of U-Net. The aim of effective exploitation of multi-scale features has brought the researcher to re-think the U-Net networks with re-designed skip connections, which we review next.

\subsection{Re-designing skip connections} \label{sec:relw:skip}
U-Net passes the features from the encoder layers to the decoder layers through skip connections and concatenates them to the upsampled outputs. The redesigning of these skip connections is ongoing research. The recent variants of U-Net like NAS-UNet \cite{weng2019unet}, U-Net3+ \cite{huang2020unet}, DC-UNet \cite{lou2021dc}, and Half-UNet \cite{lu2022half} re-design skip connections in U-Net while reducing the number of network parameters. U-Net++ and U-Net3+ change the terminology of ``skip connections'' to ``plain skip connections'' with the realisation of new skip connections. U-Net++ \cite{zhou2019unet++} propose dense nested skip connections for performance gain, but with added complexity and network parameters. U-Net3+ on the other hand re-designs the skip connections with fewer network parameters with a newly proposed full-scale skip connection. This connection utilises inter-encoder-decoder and intra-decoder skip connections to capture both fine- and coarse-grained semantics from full scales. To concatenate the maps of five scales, a multi-scale feature aggregation mechanism is developed for U-Net3+. Moving away from dense and full-skip connections, MultiResUnet \cite{ibtehaz2020multiresunet} and DR-Unet \cite{le2021dr} replace plain skip connections with \textit{Res} paths for improved performance in U-Net and ResUnet. DPN-Unets \cite{xu2021dual} utilise a dual path network (DPN) \cite{chen2017dual} to combine DenseNet and ResNet in parallel. The recent-most network that re-designed the skip connections in U-Net is UC-TransNet \cite{wang2022uctransnet}. The authors of UC-TransNet have highlighted that not all skip connections provide performance gain, and some even influence the original U-Net negatively. Their skip connection utilises a channel-wise cross-fusion transformer and channel-wise cross-attention for increased performance. It can be seen that the baseline for all the re-designing of skip connections are U-Net and ResUnet. Leaving the added complexity of transformers are other components behind, we point out that the advancement in skip connections to utilise multi-scale features is an evolving work. 

Exploiting multi-scale features with new aggregation and fusion techniques is popular in urban feature extraction from EO images \cite{wu2018automatic,wei2019toward,ji2019scale}. U-Net and U-Net3+ rely on concatenation to aggregate the multi-scale feature maps. Some networks like Half-UNet \cite{lu2022half} aggregate them using an addition operation inspired by ResNet to increase the amount of information without increasing the dimension. The final segmented output is generated without the large-scale decoder layers resulting in 100x lower network parameters with similar performance when compared to U-Net3+. The experiments also show the lower performance of U-Net3+ against U-Net in some of the datasets. In a similar attempt to subtract the components of U-Net, Fu et al. \cite{fu2021keep} design ``additive'' and ``subtractive'' variants of U-Net. The additive variants add a dense block, residual block, side-output block, and dilated convolution block to the U-Net and the subtractive variants: U-Net without Rectified Linear Units (ReLU) activation, U-Net without skip connections, and U-Net with one convolutional layer per level. The subtractive variants show a lower dice score compared to the additive variants. In their experiments, U-Net is less competitive without skip connections. Other studies have highlighted that not all skip connections between different scales of encoder-decoder improve these U-Net configurations \cite{wang2022uctransnet}. However, there is a lack of conclusions on which scales are to be focused. In this study, we take the vanilla version of U-Net, ResUnet, and U-Net3+ and enrich their skip connections. Furthermore, we confirm which scales of encoder-decoder configurations are to be focused on for performance gain.

\section{Method} \label{sec:method}
\subsection{DSCM for U-Net} \label{sec:meth:arch}
\begin{figure}[!ht]
  \centering
  \includegraphics[width=\linewidth]{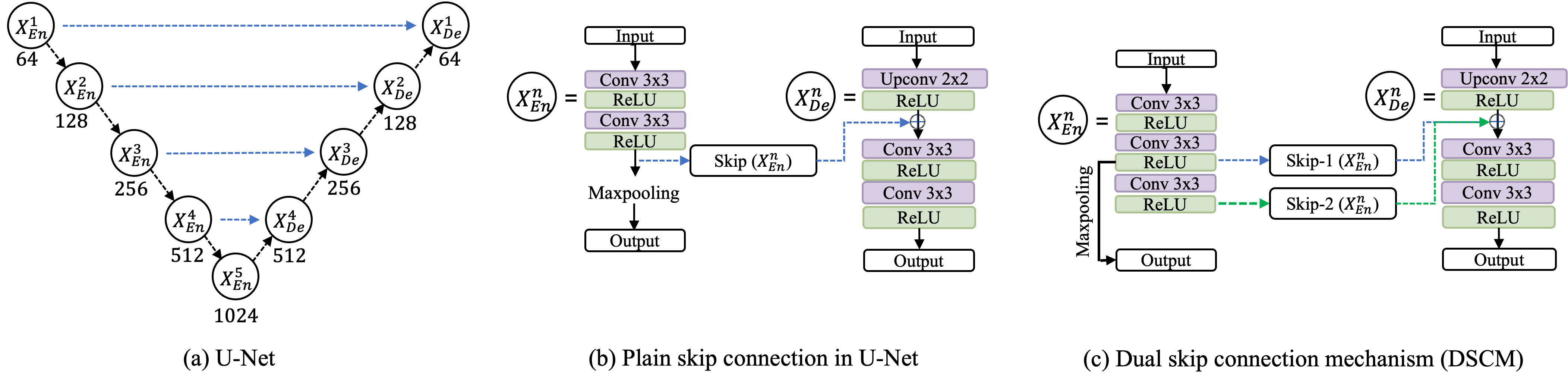}
  \caption{Illustration of the plain skip connections in U-Net and the proposed DSCM.}
      \label{fig:dualskipconnection}
      \end{figure}

Let us start with simplifying a U-Net of 5-scale layers. The encoder $En$ consecutively passes the learned feature maps along the encoder layers $X_{En}^{n}$ (where $n=1,\dots,5$). An $n^{th}$ layer of encoder $X_{En}^{n}$ consists of two recurring unpadded 3x3 convolutions $\mathcal{C}(\cdot)$ followed by a batch normalization (BN). Each $\mathcal{C}(\cdot)$ is activated with a ReLU activation function $\sigma(\cdot)$. The features are down-scaled with operation $\mathcal{D}(\cdot)$ of max-pooling with stride 2 along the consecutive encoder layers. After the final encoder layer $X_{En}^{5}$, the decoder starts by upsampling the low-resolution feature maps of $X_{En}^{5}$, and keeps doing so along the decoder layers $X_{De}^{n}$ (where $n=4,\dots,1$). An $n^{th}$ layer of decoder $X_{De}^{n}$ starts with an up-sampling operation $\mathcal{U}(\cdot)$ of a 2x2 ``up-convolution'' that halves the number of feature channels from the previous layer. The output of $\mathcal{U}(\cdot)$ is concatenated to the feature map of the corresponding encoder layer $X_{En}^{n}$, followed by a $\mathcal{C}(\cdot)$. The feature map of $X_{En}^{n}$ is brought to $X_{Dn}^{n}$ by a plain skip connection for matching index $n$. A 1x1 convolution in the final layer maps the 64-component feature vectors to the number of classes. This general 5-scale U-Net is illustrated in Figure \ref{fig:dualskipconnection}(a). With this process, U-Net utilises both low- and high-resolution features, conserving the spatial integrity of objects that is crucial in the semantic segmentation of features in EO data. However, the problem with this architecture of U-Net lies in the down-scaling as $\mathcal{D}(\cdot)$ increases the scale while halving the dimension of the feature maps, thus losing the context information at each $\mathcal{D}(\cdot)$ operation. 

To explain the proposed DSCM, let us denote one sequence of $\mathcal{C}(\cdot)$ and BN as $\mathcal{G}(\cdot)$. Then a layer of $X_{En}^{n}$ in U-Net sequentially consists of $\mathcal{C}(\cdot)$, $\mathcal{G}(\cdot)$, and $\mathcal{D}(\cdot)$ operations. To compensate for the lost context information in U-Net, we propose DSCM as illustrated in Figure \ref{fig:dualskipconnection}. With DSCM, instead of one, an encoder layer now consists of two sequential $\mathcal{G}(\cdot)$: $\mathcal{G}_{1}(\cdot)$ and $\mathcal{G}_{2}(\cdot)$. Then, an $X_{En}^{n}$ consists of $\mathcal{C}(\cdot)$, $\mathcal{G}_{1}(\cdot)$, $\mathcal{G}_{2}(\cdot)$, and $\mathcal{D}(\cdot)$ sequential operations. Similar to the U-Net, $\mathcal{D}(\cdot)$ down-scales the feature maps of the first sequence $\mathcal{G}_{1}(\cdot)$, along $X_{En}^{n}$ (where $n=1,\dots,5$). Instead of one plain skip connection, DSCM consists of two skip connections: $\mathcal{S}_{1}(\cdot)$ and $\mathcal{S}_{2}(\cdot)$, which respectively pass the features of $\mathcal{G}_{1}(\cdot)$ and $\mathcal{G}_{2}(\cdot)$ of $X_{En}^{n}$ to $X_{De}^{n}$ for matching $n$. In $X_{De}^{n}$, the feature maps brought by $\mathcal{S}_{1}(\cdot)$ and $\mathcal{S}_{2}(\cdot)$ are concatenated to the output of $\mathcal{U}(\cdot)$ that upsamples the features of the previous layer of decoder $X_{De}^{n+1}$. In case of $X_{De}^{4}$, it receives an upsampled output of $X_{En}^{5}$. The output feature map of a $X_{De}^{n}$ for $n=4,\dots,1$ can be denoted as

\begin{equation}
X_{De}^{n}=\mathcal{C}(\mathcal{C}(\mathcal{U}(X_{De}^{n+1})))
\oplus \mathcal{S}_{1}(X_{En}^{n})
\oplus \mathcal{S}_{2}(X_{En}^{n})
\label{eqn:DSCM}
\end{equation}

where, the $\mathcal{C}(\cdot)$ is supported by a ReLU $\sigma(\cdot)$, and the feature maps carried by skip connections $\mathcal{S}_{1}$ and $\mathcal{S}_{2}$ is denoted as

\begin{equation}
\mathcal{S}_{1}(X_{En}^{n})=\mathcal{G}_1(\mathcal{C}(X_{En}^{n-1}))
\label{eqn:DSCM-skip1}
\end{equation}

\begin{equation}
\mathcal{S}_{2}(X_{En}^{n})=\mathcal{G}_2(\mathcal{S}_{1}(X_{En}^{n}))
\label{eqn:DSCM-skip2}
\end{equation}

\subsection{DSCM in different scale features of U-Net} \label{sec:meth:dualskip}
The DSCM can be plugged into any scale layers of U-Net. To study the effects of DSCM on different scales of fine-grained detailed information and coarse-grained semantic information, we propose three novel U-Net network architectures: dual skip on large-scale features (DS-UNet-L), dual skip on small-scale features (DS-UNet-S), and dual skip on all scale features (DS-UNet-A). 

\begin{enumerate}
    \setcounter{enumi}{0}
    \item \textit{DS-UNet-L} applies DSCM between the large-scale layers of encoder $X_{En}^{n}$ and decoder $X_{De}^{n}$ for $n\in [3,4]$.
    \item \textit{DS-UNet-S} applies DSCM between the small-scale layers of encoder $X_{En}^{n}$ and decoder $X_{De}^{n}$ for $n\in [1,2]$.
    \item \textit{DS-UNet-A} applies DSCM between all four scale layers of encoder $X_{En}^{n}$ and decoder $X_{De}^{n}$ for $n\in [1,2,3,4]$.
\end{enumerate}

\subsection{DRSCM for ResUnet} \label{sec:meth:dsresunet}
\begin{figure}[!ht]
  \centering
  \includegraphics[width=\linewidth]{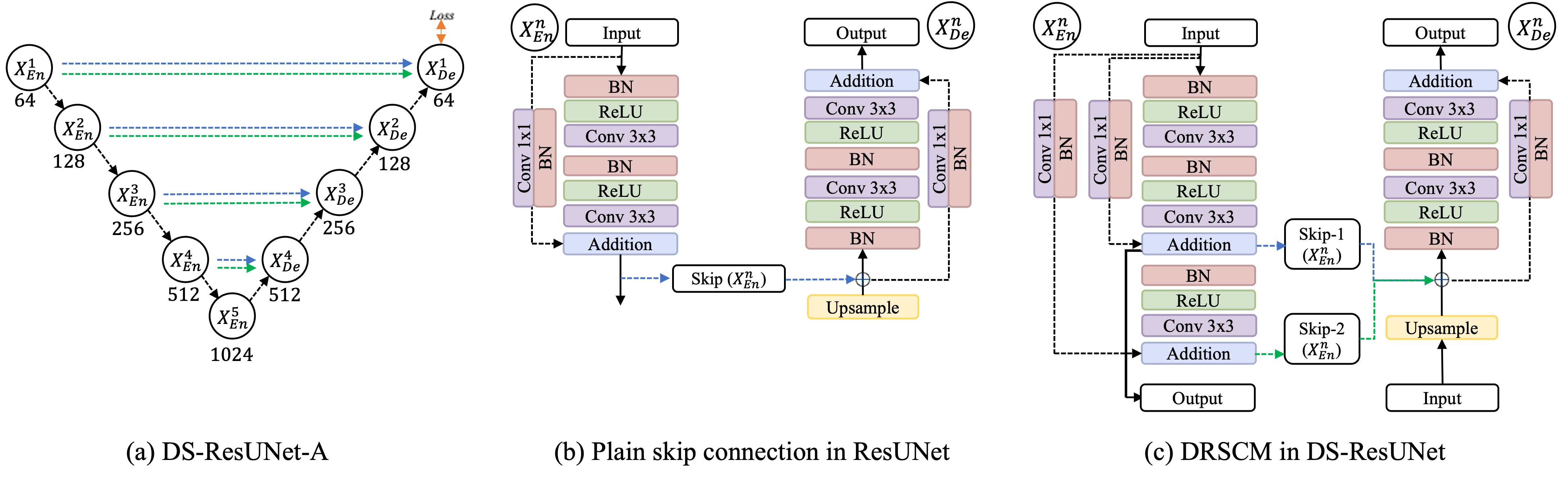}
  \caption{Illustration of DS-ResUNet-A along with the differences between the plain skip connections in ResUnet and the proposed DRSCM}
      \label{fig:dsresunet}
      \end{figure}

DSCM cannot be directly implemented in ResUnet because of its residual units. For this, we propose a DRSCM for ResUnet. We name the ResUnet with DRSCM as a DS-ResUnet. In addition to the U-Net configuration, a ResUnet includes a skip connection (we name it respath) within its residual unit \cite{zhang2018road}. Therefore, to deepen the residual unit with one more 3x3 convolution, one more respath needs to be added as shown in the illustration of the proposed DRSCM in Figure \ref{fig:dsresunet}(c). Unlike the existing dual skips in \cite{le2021dr} and \cite{xu2021dual}, the residual unit is deeper and denser in our proposed DRSCM with an added 3x3 convolution layer $\mathcal{C}(\cdot)$, BN, ReLU $\sigma(\cdot)$, and a respath $\mathcal{R}(\cdot)$. 

Let us start by explaining a ResUNet of 5-scale layers. Encoder $En$ consecutively passes the learned feature maps along $X_{En}^{n}$ (where $n=1,\dots,5$). Lets denote one sequence of BN and unpadded 3x3 convolution $\mathcal{C}(\cdot)$ as $\mathcal{G}(\cdot)$. Each $\mathcal{C}(\cdot)$ is activated with a ReLU $\sigma(\cdot)$. Then, $X_{En}^{n}$ consists of two sequences of $\mathcal{G}(\cdot)$ i.e., $\mathcal{G}_{1}(\cdot)$ followed by $\mathcal{G}_{2}(\cdot)$. The output of $\mathcal{G}_{2}(\cdot)$ is added to the output of a respath skip connection $\mathcal{R}(\cdot)$. A respath skip connection $\mathcal{R}(\cdot)$ consists of a 1x1 $\mathcal{C}(\cdot)$ followed by a BN on the input feature map of $X_{En}^{n-1}$. 

With DRSCM integrated into a residual block of a ResUNet, we add a third $\mathcal{G}(\cdot)$ and $\mathcal{R}(\cdot)$ operations to deepen the residual block. Now, a $X_{En}^{n}$ consists of $\mathcal{G}_{1}(\cdot)$, $\mathcal{G}_{2}(\cdot)$, $\mathcal{R}_{1}(\cdot)$, $\mathcal{G}_{3}(\cdot)$, and $\mathcal{R}_{2}(\cdot)$ sequential operations. The output of $\mathcal{G}_{3}(\cdot)$ is added to the output of second respath skip connection $\mathcal{R}_{2}(\cdot)$. The output features of the $\mathcal{R}_{1}(\cdot)$ are down-scaled with operation $\mathcal{D}(\cdot)$ of max-pooling with stride 2 along $n$ residual units. After the final encoder layer $X_{En}^{5}$, the decoder starts by up-sampling the low-resolution feature maps of $X_{En}^{5}$, and keeps on doing so along the decoder layers $X_{De}^{n}$ (where $n=4,\dots,1$). An $n_{th}$ decoder layer $X_{De}^{n}$ starts with an up-sampling operation $\mathcal{U}(\cdot)$ of 2x2 ``up-convolution''. In case of $X_{De}^{4}$, it receives an upsampled output of $X_{En}^{5}$. The output of $\mathcal{U}(\cdot)$ is concatenated to the two sets of feature maps of the corresponding encoder layer $X_{En}^{n}$, followed by two consecutive $\mathcal{G}(\cdot)$. The two sets of feature maps are brought from $X_{En}^{n}$ to $X_{De}^{n}$ for matching index $n$ by two skip connections of DRSCM, each consisting of the output of $\mathcal{R}_{1}(\cdot)$+$\mathcal{G}_{2}(\cdot)$ and $\mathcal{R}_{2}(\cdot)$+$\mathcal{G}_{3}(\cdot)$. The output feature map of a $X_{De}^{n}$ for $n=4,\dots,1$ can be denoted as

\begin{equation}
\begin{split}
X_{De}^{n}=[\mathcal{G}(\mathcal{G}(\mathcal{U}(X_{De}^{n+1})))
\oplus \mathcal{S}_{1}(X_{En}^{n})
\oplus \mathcal{S}_{2}(X_{En}^{n})] \\ + \mathcal{R}(X_{De}^{n+1})
\label{eqn:DRSCM}
\end{split}
\end{equation}

where, the feature maps carried by skip connections $\mathcal{S}_{1}$ and $\mathcal{S}_{2}$ can be denoted as

\begin{equation}
\mathcal{S}_{1}(X_{En}^{n})=\mathcal{G}_{2}(\mathcal{G}_{1}(X_{En}^{n-1}))+\mathcal{R}_{1}(X_{En}^{n-1})
\label{eqn:DRSCM-skip1}
\end{equation}

\begin{equation}
\mathcal{S}_{2}(X_{En}^{n})=\mathcal{G}_{3}(\mathcal{S}_{1}(X_{En}^{n}))+\mathcal{R}_{2}(X_{En}^{n-1})
\label{eqn:DRSCM-skip2}
\end{equation}

Similar to the three variations of DS-UNet, we experiment with the integration of DRSCM in the large-scale (DS-ResUNet-L), small-scale (DS-ResUNet-S), and all-scale (DS-ResUNet-A) features.
      
\subsection{DFSCM for U-Net3+} \label{sec:meth:dsunet3+}
The integration of dual skip connections in U-Net3+ is more sophisticated because of the existing full-skip connections in U-Net3+ \cite{huang2020unet}. Unlike the plain skip connections of U-Net and ResUnet that pass the features from $X_{En}^{n}$ to $X_{De}^{n}$ only for matching index $n$, the full-skip connections allow aggregating the feature maps of all $n$ scales to capture both fine- and coarse-grained semantics in $X_{De}^{n}$. Let us take an example of the third decoder layer $X_{De}^{3}$ of U-Net3+. $X_{De}^{3}$ receives the feature maps of (i) same-scale encoder layer $X_{En}^{3}$ similar to U-Net, (ii) smaller-scale encoder layer $X_{En}^{2}$ and $X_{En}^{1}$ carrying coarse-grained detailed information through inter encoder-decoder skip connections supported by non-overlapping max pooling operations, and (iii) larger-scale decoder layer $X_{De}^{4}$ and $X_{De}^{5}$ carrying fine-grained semantic information through intra-decoder connections supported by bilinear interpolation. The number of channels in all incoming five same-resolution feature maps is unified with 64 filters of 3x3 size. A feature aggregation mechanism is applied on the concatenated maps of five scales that consist of 320 (64 times 5) filters of 3x3 size, a BN, and a ReLU activation function. The feature aggregation mechanism in U-Net3+ assumes equal weights for all feature maps at different scales. However, the different scales of feature maps possess different levels of discrimination. The context information is more intact in the small-scale feature maps as the large-scale features get smaller in size and lose them because of the downsampling operations.

\begin{figure}[!hbt]
  \centering
  \includegraphics[width=\linewidth]{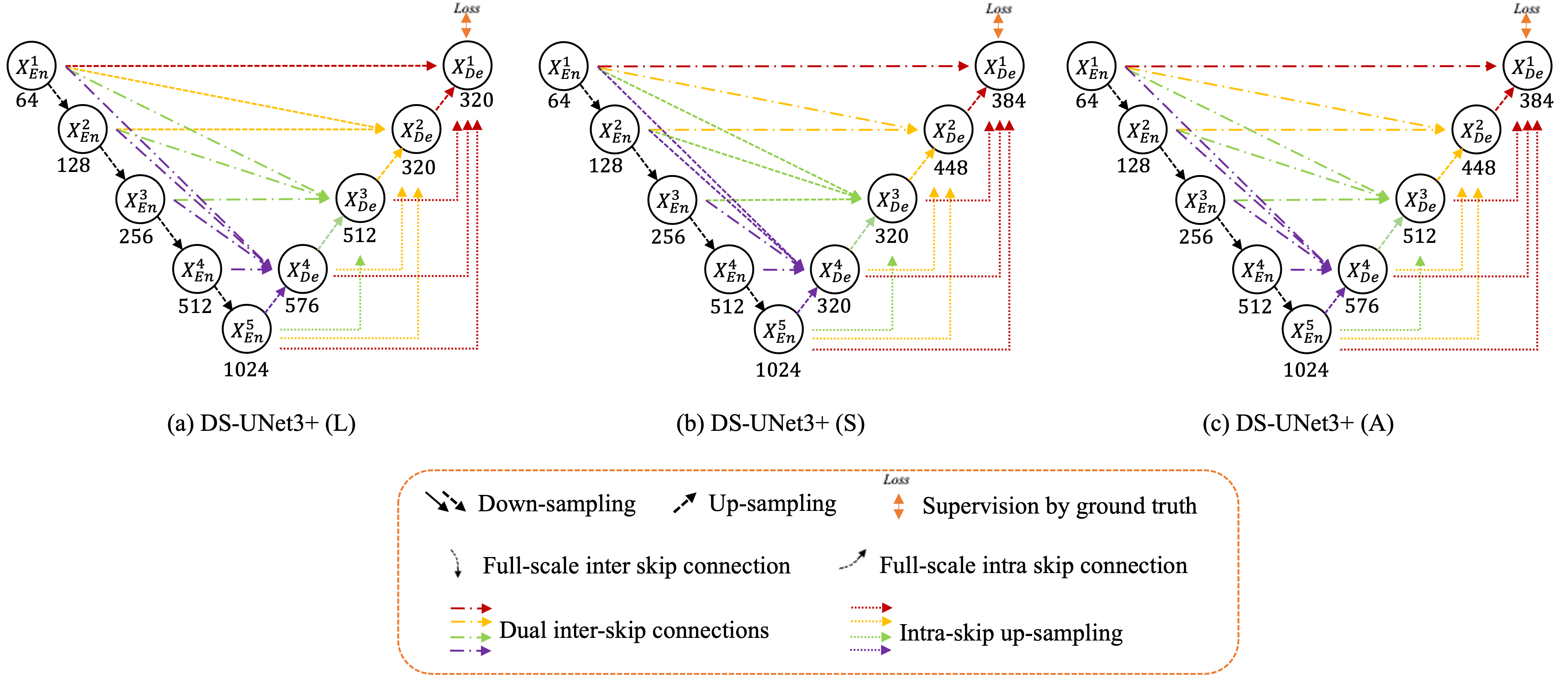}
  \caption{Illustration of the proposed DS-UNet3+ networks DS-UNet3+(L), DS-UNet3+(S), and DS-UNet3+(A).}
      \label{fig:dsunet3plus}
      \end{figure}

\begin{figure}[!hbt]
  \centering
  \includegraphics[width=0.8\linewidth]{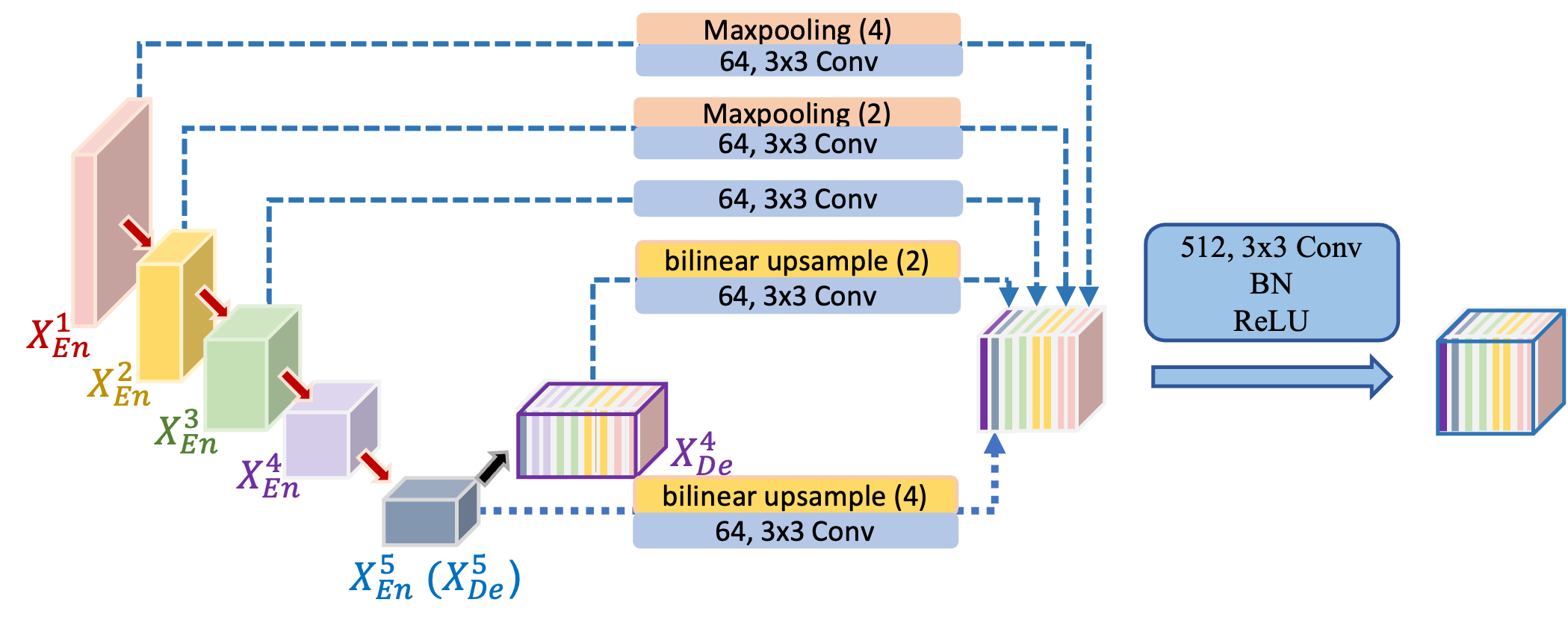}
  \caption{Illustration of dual skip feature aggregation mechanism (abbr. DSFAM) at the third decoder layer $X_{De}^{3}$ of DS-UNet3+ (figure adopted and modified from \cite{huang2020unet}).}
      \label{fig:dsunet3plus-De3}
      \end{figure}

We integrate the concepts of DSCM and full-scale skip connections from U-Net3+ to propose DFSCM for U-Net3+. We name the formulated networks DS-UNet3+. The difference in U-Net3+ and DS-UNet3+ is illustrated in Figure \ref{fig:dsunet3plus}. The skip connections are similar to those of U-Net3+, except the single inter-encoder-decoder plain skip connections are all replaced by two plain skip connections of DSCM. To support the two connections in DS-UNet3+, we develop a dual skip feature aggregation mechanism (abbr. DSFAM). The DSFAM consists of a different number of filters when compared to U-Net3+. For example, Figure \ref{fig:dsunet3plus-De3} illustrates the construction of feature maps in the third decoder layer of DS-UNet3+. 

The $X_{De}^{3}$ receives the feature maps of same-scale encoder layer $X_{En}^{3}$ with two plain skip connections. The feature maps of smaller-scale encoder layer $X_{En}^{2}$ and $X_{En}^{1}$ carrying coarse-grained detailed information are transported through two inter-encoder-decoder skip connections supported by down-scaling operation $\mathcal{D}(\cdot)$ of non-overlapping max pooling operations. Finally, the feature maps from larger-scale decoder layer $X_{De}^{4}$ and $X_{De}^{5}$ carrying fine-grained semantic information are transported through intra-decoder connections supported by upsampling operation $\mathcal{U}(\cdot)$ of bilinear interpolation. In case of $X_{De}^{4}$, it receives an upsampled output of $X_{En}^{5}$. Similar to U-Net3+, the number of channels in all incoming same-resolution feature maps is unified with 64 filters of 3x3 size. The output feature map $f$ of a $X_{De}^{n}$ for $n=4,\dots,1$ of a $N$-scaled DS-UNet3+ can be denoted as

\begin{equation}
\begin{split}
X_{De}^{n}=\mathcal{C} \big( \mathcal{C} \big( \mathcal{U}(X_{De}^{n+1}) \big) \big)  
\oplus 
\underbrace{ \mathcal{DS}_{F} \big( X_{En}^{n}, \dots, \mathcal{D}_{2i}(X_{En}^{1})\big) }_{\text{Scales}: \ n^{th} \sim 1^{th}} 
\\
\oplus  
\underbrace{ \mathcal{S}_{De} \big( \mathcal{U}_{2j}(X_{De}^{n}), \dots, \mathcal{U}_{2j}(X_{De}^{N}) \big) }_{\text{Scales}: \ (n+1)^{th} \sim (N-1)^{th}},
\\
i = k \ (\text{for} \ k=1 \ \text{to} \ n),
\\
j = m \ (\text{for} \ m=1 \ \text{to} \ n-k+1)
\label{eqn:DSFAM}
\end{split}
\end{equation}

where, $\mathcal{C}(\cdot)$ denotes unpadded 3x3 convolution activated with ReLU $\sigma(\cdot)$. $\mathcal{DS}_{F}(\cdot)$ denotes DFSCM that brings the features from $\mathcal{S}_{1}$ (refer to Eqn. \ref{eqn:DSCM-skip1}) and $\mathcal{S}_{2}$ (refer to Eqn. \ref{eqn:DSCM-skip2}) from each encoder layers $X_{En}^{n}, \dots, X_{En}^{1}$. Similar to the DS-UNets, the second and third convolutional layer in each encoder block is followed by a BN. $\mathcal{S}_{De}(\cdot)$ refers to the intra-decoder connections that bring the features from previous decoder layers $X_{De}^{n}, \dots, X_{De}^{N}$. The value of $i$ and $j$ denote stride size for down-sampling operation $\mathcal{D}_{2i}(\cdot)$ and bilinear up-sampling operation $\mathcal{U}_{2j}(\cdot)$ respectively. Here, $i$ ranges from $1$ to $n$ and $j$ ranges from $1$ to $n-i+1$ depending upon the value of $n$.

Similar to DS-UNet and DS-ResUNet, three variants of DS-UNet3+ are studied: DS-UNet3+(L), DS-UNet3+(S), and DS-UNet3+(A). In DS-UNet3+(A), the DSFAM concatenates 9, 8, 7, and 6 sets of feature maps with a total of 576, 512, 448, and 384 filters of 3x3 size on decoder layers $X_{De}^{4}$, $X_{De}^{3}$, $X_{De}^{2}$, and $X_{De}^{1}$ respectively. In DS-UNet3+(L), the DFSCM is integrated between third ($X_{En}^{3}$ to $X_{De}^{3}$) and fourth ($X_{En}^{4}$ to $X_{De}^{4}$) scale layers of U-Net3+. Similarly, in the small-scale variation DS-UNet3+(S), the DFSCM is integrated between first ($X_{En}^{1}$ to $X_{De}^{1}$) and second ($X_{En}^{2}$ to $X_{De}^{2}$) scale layers. The proposed DS-UNet3+ networks consist of fewer network parameters than U-Net, ResUnet, DS-UNets, and DS-ResUNets.

\section{Dataset and Experimental Setup} \label{sec:experiments}

\subsection{Datasets} \label{sec:exp:data}
We experiment on a newly developed multi-resolution building footprint dataset and three existing datasets of VHR and the high-resolution type.

The first dataset that we have developed (label, image) is named as Melbourne building footprint dataset (abbr. MELB). It is the first multi-resolution building dataset that covers samples from a complex urban environment i.e., the City of Melbourne, Australia. The labels are developed by masking and tiling the building roof samples provided by the City of Melbourne. The corresponding image tiles of 0.3m, 0.6m, and 1.2m spatial resolution of the labels are collected from Nearmap's API service. The number of training and validation samples are divided as 70\% and 30\% respectively. Therefore, this dataset is prepared in an end-to-end manner, without manual annotations. It is made sure that the image and labels are of the same projection system and of the same year.

The second dataset is a 1.2m subset of the MELB dataset that we developed in our previous work \cite{neupane2022building}. We use this dataset to experiment with the proposed networks on high-resolution images.

The third dataset is the high-resolution (1m) Massachusetts Building dataset \cite{mnih2013machine}. The original 1500x1500 tiles are cropped to 256x256 by generating a grid of coordinates. The validation images provided are used for validation. The partial tiles on the edges of the tiles are ignored, by iterating only through the Cartesian product between the two intervals $range(0, h-h \bmod d, d)$ and $range(0,w-w \bmod d, d)$ for width $w$, height $h$, and output tile size of 256 as $d$. Let $H=h-h \bmod d$ and $W=w-w \bmod d$. Then the Cartesian product of the two intervals is the set of all ordered pairs $(i,j)$, and can be defined as

\begin{equation}
\begin{split}
\mathcal{P} = \{(i,j)\ |\ 0\le i<H, 0\le j<W, i\in\mathbb{N}, j\in\mathbb{N}, \\  i\equiv 0\ (\textrm{mod}\ d), j\equiv 0\ (\textrm{mod}\ d)\}
\label{eqn:cartesian}
\end{split}
\end{equation}

where, $i$, and $j$ are non-negative integers among the set of natural numbers $\mathbb{N}$.

The fourth dataset is the VHR WHU Building dataset \cite{ji2018fully} (abbr. WHU). It includes satellite images of Christchurch, New Zealand with a spatial resolution of 0.3m. To maintain uniform image size between all the datasets in our experiments, the originally 512x512-sized image patches of the WHU are tiled into 256x256, increasing the number of image-label pairs by four times. Thus, 23088 training tiles and 9664 validation tiles are prepared.

\subsection{Training details} \label{sec:exp:traindetails}
The proposed variations of DS-UNet, DS-ResUNet, and DS-UNet3+ are compared to the vanilla U-Net, ResUnet, and U-Net3+. Furthermore, the networks are also compared to U-Net++ \cite{zhou2019unet++}, TransUnet \cite{chen2021transunet}, Swin-Unet \cite{cao2022swin}, Attention U-Net \cite{oktay2018attention}, U$^2$Net \cite{qin2020u2}, FCN8s \cite{long2015fully}, DeepLabv3+ \cite{chen2017deeplab}, and SegNet \cite{badrinarayanan2017segnet}. For uniformity in experiments, the same loss function is used to evaluate all networks. A hybrid loss function that is the sum of the binary focal loss (BFL) \cite{lin2017focal} and dice loss \cite{sudre2017generalised} is used as the loss function to monitor the proposed and the compared networks in our experiments. BFL is a variation of the standard Cross-Entropy (CE) loss that adds a focusing mechanism to reduce the loss contribution of easy-to-classify examples, putting more emphasis on hard-to-classify examples, which is particularly useful for imbalanced datasets. Dice loss calculates the measure of overlap to assess the performance of segmentation when a ground truth (GT) is available. BFL and Dice loss are denoted by Eqn. (\ref{eqn:bfl}) and (\ref{eqn:diceloss}) as

    \begin{equation}\label{eqn:bfl}
        L_{BFL}(y,\hat{p}) = - y \alpha (1 - \hat{p})^\gamma \log(\hat{p}) - (1 - y) \alpha \hat{p}^\gamma \log(1 - \hat{p})
    \end{equation}

    \begin{equation}\label{eqn:diceloss}
        L_{dice}(y,\hat{p}) = 1 - \frac{2y\hat{p}+1} {y+\hat{p}+1} 
    \end{equation}
    
where $y$ and $\hat{p}$ represent the GT and prediction respectively. In BFL, $\alpha$ is the weighing factor introduced to address the class imbalance. Similarly, $\gamma$ is a focusing parameter such that when $\gamma = 0$, BFL is equivalent to CE, and as $\gamma$ is increased the effect of the modulating factor is likewise increased. In dice loss, the smooth value of 1 is added in the numerator and denominator, considering the edge case scenario of $y=\hat{p}=0$. The product $y\hat{p}$ represents the intersection between the GT and prediction.

All DL networks are wrapped in the Keras framework using \textit{Segmentation Models library} \cite{Yakubovskiy2019}. The step/epoch is set as the ratio of the number of training images to the batch size of 16. The epoch is set such that the total number of steps is kept the same or similar (approx. 60000). A learning rate of 1e-4 is used to train all the networks. The learning rate is reduced upon a plateau of the loss by a factor of 0.1 and patience of 10 epochs and stopped at the saturation of loss in the rate of 1e-6. \textit{RMSProp}, \textit{He Normal}, and ReLU are the optimizer, initializer, and activation functions respectively. A \textit{sigmoid} function is used to obtain the final output maps as the dataset is binary, and a \textit{dropout} of 50\% is used to avoid over-fitting. The hyper-parameters are kept the same to train all the proposed, and SOTA networks. All networks are trained without deep supervision. 

\subsection{Evaluation Metrics} \label{sec:exp:eval}
Four accuracy measures are used for evaluation in this work: precision (P), recall (R), intersection over union, and F1 score. P and R measure the accuracy of positive predictions made by the segmentation model, and the completeness of the segmentation results. IoU and F1 are calculated from the `area-of-overlap' between prediction and binary labels and the `area-of-union' (all of the predictions + binary labels - the overlap). The network with the highest F1 score is chosen as the network to study the knowledge transfer techniques. The mathematical notation of the four measures are

    \begin{equation}\label{eqn:sensitivity}
        \text{P} = \frac{TP} {TP + FP}
    \end{equation}
    
    \begin{equation}\label{eqn:specificity}
        \text{R} = 1 - \frac{TP} {TP + FN}
    \end{equation}

    \begin{equation}\label{eqn:iou}
        \text{IoU} = \frac{TP} {TP + FN + FP}
    \end{equation}

    \begin{equation}\label{eqn:f1score}
        \text{F1 score} = \frac{2 \times TP} {2 \times TP + FN + FP} 
    \end{equation}

where, TP implies prediction = 1, label = 1; TN implies prediction = 0, label = 0; FP implies prediction = 1, label = 0; and FN implies prediction = 0, label = 1.

\section{Results and Discussion} \label{sec:res}
The results and comparisons from the experiments are presented for four experimental settings categorised based on spatial resolutions: multi-resolution, high-resolution, and VHR. The proposed networks are compared to the vanilla versions of U-Net, ResUnet, U-Net3+, and other SOTA networks. Further, the DSCM is studied at different scales of U-Net and differently scaled U-Nets as an ablation study. Overall, 90 experiments are performed in this section to evaluate the proposed dual skip connection mechanisms in U-Net, ResUnet, and U-Net3+. The limitations of the experiments finalise the discussion at the end of this section.

\subsection{Results on multi-resolution dataset (MELB)} \label{sec:res:MELB}
\begin{figure}[!ht]
  \centering
  \includegraphics[width=\linewidth]{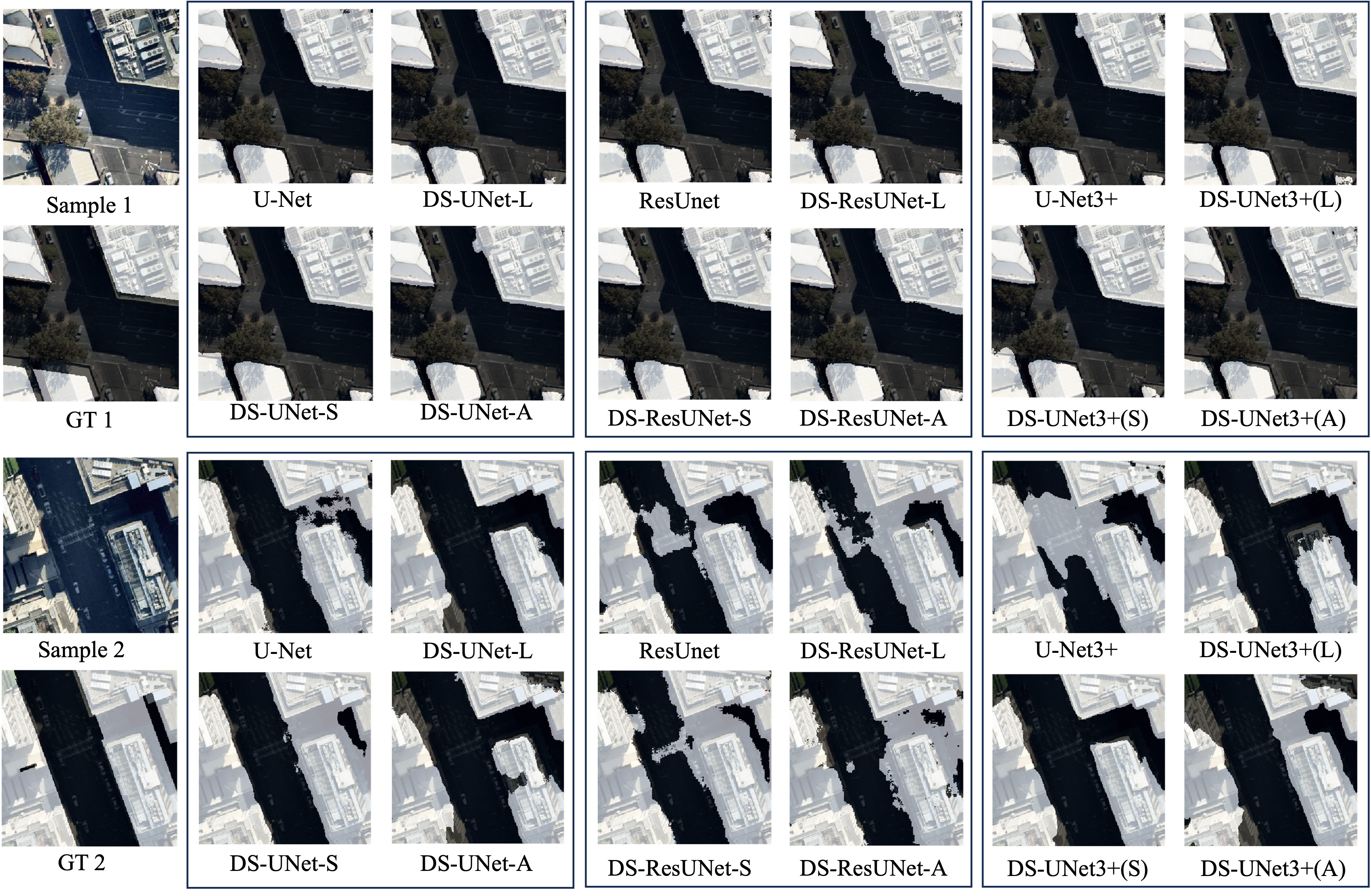}
  \caption{Segmentation output from the proposed networks and their original vanilla baseline networks on the MELB building dataset.}
      \label{fig:MELB-results}
      \end{figure}

Table \ref{tab:res:MELB} presents the results of the proposed networks, their vanilla baseline architectures, and SOTA networks on the MELB dataset. Figure \ref{fig:MELB-results} shows the sample results from all the proposed and original vanilla networks on the MELB dataset with two samples of high-rise buildings with complex roof structures. The results on this dataset can be divided into the proposed variants of the three networks and the SOTA comparison:

\begin{itemize}
    \item \textit{DSCM for U-Net (DS-UNets)}: All three versions of the proposed DS-UNet outperform U-Net on the MELB dataset in terms of P, IoU and F1. DS-UNet-L shows the highest IoU and F1 among the three versions and also outperforms all compared SOTA networks except U$^2$Net. The results on the VHR dataset show that deepening the large-scale features in U-Net with DSCM results in the highest performance with about a 14\% increase in network parameters. The small-scale features can be kept the same without being doubled as seen from DS-UNet-A. This supports our initial argument that the trade-off between the use of context and precise location can be improved in U-Net. This can be done by increasing the large-scale feature maps, where the contexts are lost due to down-sampling operations such as max-pooling.

    \item \textit{DRSCM for ResUnet (DS-ResUNet)}: The DS-ResUNet networks are unable to outperform their vanilla ResUnet baseline in terms of R, IoU and F1. DS-ResUNet-L produced the highest P and among the three proposed variants, DS-ResUNet-S produced the highest IoU and F1.

    \item \textit{DFSCM for U-Net3+ (DS-UNet3+)}: The results from DFSCM on DS-UNet3+ also report success in the MELB dataset. All three versions significantly outperform U-Net3+ in the majority of evaluation measures. DS-UNet3+(S) yields the highest scores in IoU and F1 with the least increase in network parameters.

    \item \textit{SOTA comparison}: The proposed dual skip mechanisms yielded performance gain only for the scale-variants of DS-UNet and DS-UNet3+. DS-UNet-L and DS-UNet3+(S) produced IoU and F1 of up to 0.837 and 0.905 respectively. These scores are higher than 10 out of 11 SOTA networks that we compare to. U$^2$Net yielded the highest IoU and F1 with around 196 million network parameters. Compared to DS-UNet3+(S), U$^2$Net has more than 8 times higher parameters. The majority of the proposed networks also outperform Transformer-based TransUnet and SwinUnet. DS-UNet3+(S) outperforms TransUnet with almost 19 times lower network parameters. FCN8s network is unable to converge the R, IoU and F1 scores in this dataset.
\end{itemize}

\begin{table}[!hbt]
\centering
\caption{Performance of DS-UNet, DS-ResUNet, and DS-UNet3+ on multi-resolution (0.3m + 0.6m + 1.2m) MELB dataset and their comparison to the SOTA networks. The highest scores are highlighted in bold in each group of experiments and `-' shows that the network failed to yield the scores. The top 3 scores are ranked in order with a postscript for each evaluation metric.}
\label{tab:res:MELB}
\centering
\small
\begin{tabular}{lccccc}
\hline
Networks & Par. (M) & P & R & IoU & F1 \\ \hline
U-Net & 31.041 & 0.682 & \textbf{1.000}$^1$ & 0.682 & 0.794 \\
DS-UNet-L & 36.943 & 0.923 & 0.896 & \textbf{0.837}$^3$ & \textbf{0.905}$^2$ \\
DS-UNet-S & 31.410 & \textbf{0.935} & 0.863 & 0.818 & 0.892 \\
DS-UNet-A & 37.312 & 0.899 & 0.913 & 0.833 & 0.901 \\ \hline
ResUnet & 75.346 & 0.911 & \textbf{0.901} & \textbf{0.833} & \textbf{0.902} \\
DS-ResUNet-L & 85.024 & \textbf{0.934} & 0.864 & 0.819 & 0.893 \\
DS-ResUNet-S & 75.951 & 0.918 & 0.892 & 0.831 & 0.901 \\
DS-ResUNet-A & 85.629 & 0.913 & 0.896 & 0.830 & 0.900 \\ \hline
U-Net3+ & 22.891 & 0.922 & 0.887 & 0.828 & 0.899 \\
DS-UNet3+(L) & 27.101 & \textbf{0.936}$^3$ & 0.875 & 0.829 & 0.899 \\
DS-UNet3+(S) & 23.335 & 0.923 & 0.897 & \textbf{0.838}$^2$ & \textbf{0.905}$^2$ \\
DS-UNet3+(A) & 27.359 & 0.883 & \textbf{0.935}$^2$ & 0.836 & 0.903$^3$ \\ \hline
U-Net++ & 34.538 & 0.930 & 0.883 & 0.833 & 0.901 \\
TransUnet & 437.277 & 0.937$^2$ & 0.879 & 0.834 & 0.903$^3$ \\
Swin-Unet & 48.174 & 0.869 & 0.911 & 0.802 & 0.882 \\
Att.Unet & 34.884 & 0.938$^1$ & 0.871 & 0.828 & 0.898 \\
U$^2$Net & 195.868 & 0.915 & 0.916$^3$ & 0.849$^1$ & 0.912$^1$ \\
FCN8s & 14.741 & - & - & - & - \\
DeepLabv3+ & 11.852 & 0.892 & 0.905 & 0.820 & 0.893 \\
SegNet & 29.458 & 0.923 & 0.878 & 0.823 & 0.896 \\ \hline
\end{tabular}
\end{table}

\subsubsection{Ablation Study of DS-UNets} \label{sec:res:MELB:ablation}
\begin{figure*}[!hbt]
  \centering
  \includegraphics[width=\linewidth]{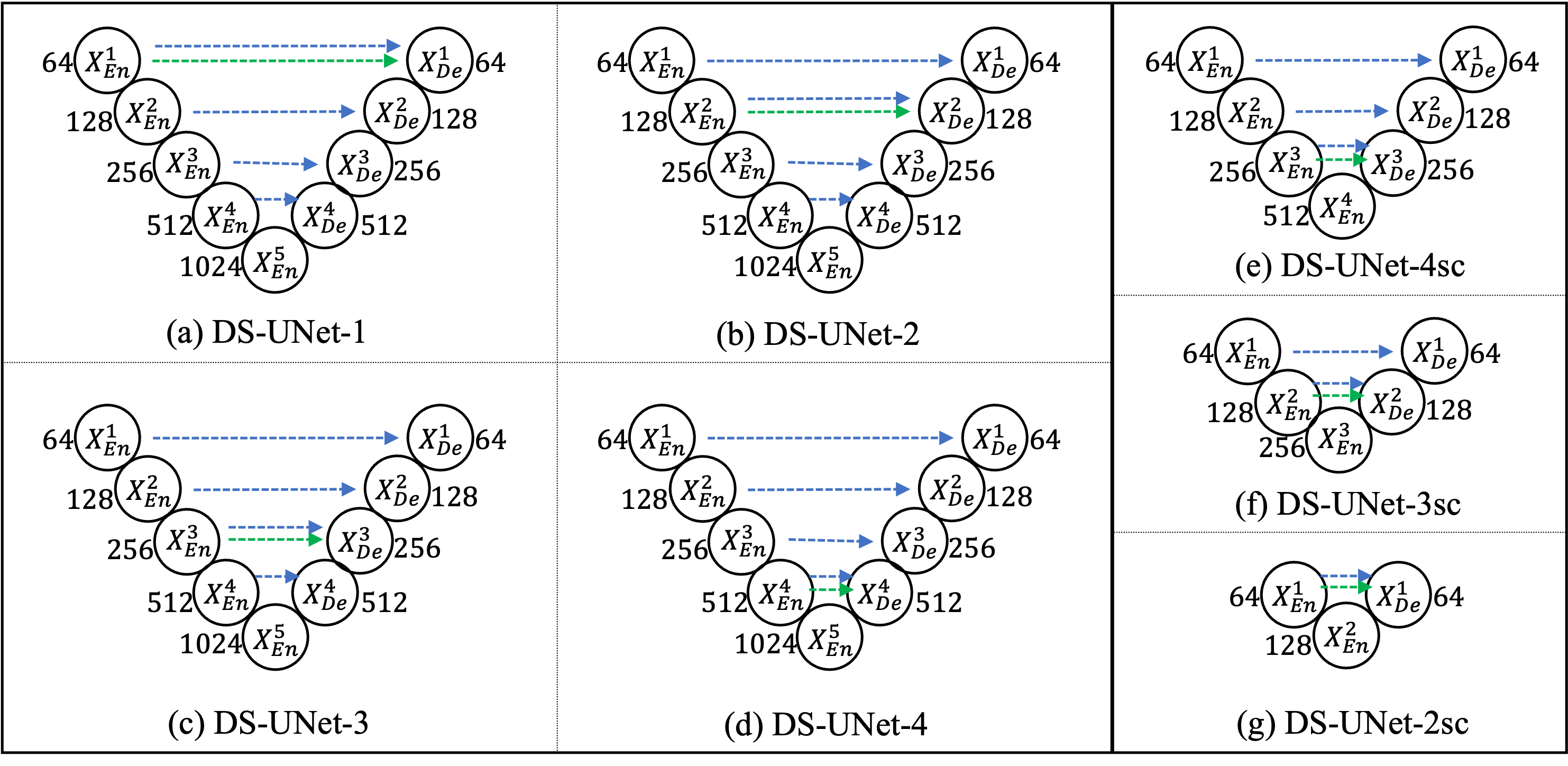}
  \caption{Illustration of the seven DS-UNet networks for ablation study. Sub-figure (a)-(d) and sub-figure (e)-(g) show the DS-UNets of the first and second group of ablation experiments respectively.}
      \label{fig:ablation}
      \end{figure*}

Results in the MELB dataset have shown that not all skip connections improve the performance of U-Net. An ablation study is conducted further to study the effectiveness of the proposed DSCM on different scale layers of U-Net and different scaled U-Nets. The design of this ablation study is categorised into two groups. The first group of ablation experiments take a U-Net of five-scale layers as shown in Figure \ref{fig:dualskipconnection}(a) before, and applies DSCM between each scale layer of its encoder $X_{En}^{n}$ and decoder $X_{De}^{n}$. Figure \ref{fig:ablation}(a)-(d) illustrates the four DS-UNet formed in this group of experiments: DS-UNet-1, DS-UNet-2, DS-UNet-3, and DS-UNet-4. The second group of ablation experiments investigates the smaller DS-UNets with a lower number of scales as illustrated in Figures \ref{fig:ablation}(e)-(g). Unlike the DS-UNets of five scale layers, these experiments compare DS-UNets of 4, 3, and 2 scale layers and are named DS-UNet-4sc, DS-UNet-3sc, and DS-UNet-2sc respectively. 

The performance evaluation of the first ablation group is shown in Table \ref{tab:res:indiscale}. All variants from DS-UNet-1 to DS-UNet-4 outperformed U-Net in terms of P, IoU and F1. DS-UNet-3 produced the highest IoU and F1 scores reporting that deepening just the third skip connection in U-Net can significantly increase the IoU and F1 by 16\% and 11.3\% respectively. This points out to the conclusion that the DS-UNet-L from Table \ref{tab:res:MELB} have inherited its highest scores from the denser third skip connection.

\begin{table}[!hbt]
\centering
\caption{First group of ablation: proposed DSCM between each scale of encoder-decoder layers, and their comparison to U-Net. The highest scores are highlighted in bold.}
\label{tab:res:indiscale}
\centering
\small
\begin{tabular}{lccccc}
\hline
Networks & Par. (M) & P & R & IoU & F1 \\ \hline
U-Net & 31.041 & 0.682 & \textbf{1.000} & 0.682 & 0.794 \\
DS-UNet-1 & 31.114 & 0.920 & 0.900 & 0.838 & 0.905 \\
DS-UNet-2 & 31.336 & \textbf{0.933} & 0.873 & 0.825 & 0.897 \\
DS-UNet-3 & 32.221 & 0.910 & 0.912 & \textbf{0.842} & \textbf{0.907} \\
DS-UNet-4 & 35.761 & 0.923 & 0.873 & 0.817 & 0.890 \\ \hline
\end{tabular}
\end{table}

Table \ref{tab:res:smallscales} presents the comparison of the three DS-UNets and their corresponding U-Net baselines from the second group of the ablation experiment. These three DS-UNets integrate DSCM on only the skip connections between the largest scale encoder-decoder layers in the network structure as DS-UNet with denser large scale layers (DS-UNet-L) yielded the highest IoU and F1 in Table \ref{tab:res:MELB}. As reported, DSCM does not improve the evaluation measures of 2-scale, 3-scale, and 4-scale U-Nets but improves those of 5-scale U-Net. 

\begin{table}[!hbt]
\centering
\caption{Second group of ablation: proposed DSCM on 4, 3, and 2 scale U-Nets. The highest scores are highlighted in bold.}
\label{tab:res:smallscales}
\centering
\small
\begin{tabular}{lccccc}
\hline
Networks & Par. (M) & P & R & IoU & F1 \\ \hline
UNet-2sc & 0.405 & 0.858 & 0.893 & 0.780 & 0.868 \\
DS-UNet-2sc & 0.479 & 0.897 & 0.837 & 0.766 & 0.859 \\ \hline
UNet-3sc & 1.865 & 0.912 & 0.882 & 0.816 & 0.891 \\
DS-UNet-3sc & 2.161 & 0.916 & 0.873 & 0.813 & 0.890 \\ \hline
UNet-4sc & 7.702 & 0.922 & 0.893 & 0.835 & 0.903 \\
DS-UNet-4sc & 8.883 & \textbf{0.940} & 0.849 & 0.810 & 0.887 \\ \hline
UNet-5sc & 31.041 & 0.682 & \textbf{1.000} & 0.682 & 0.794 \\
DS-UNet-5sc & 35.761 & 0.923 & 0.896 & \textbf{0.837} & \textbf{0.905} \\ \hline
\end{tabular}
\end{table}

\subsection{Results on 1.2m subset of MELB dataset} \label{sec:res:MELB12}
Table \ref{tab:res:melb12} presents the results from the networks on the 1.2m high-resolution building footprint dataset, which is a subset of MELB. DS-UNet variants are unable to outperform U-Net in this subset in R, IoU and F1. DS-UNet-S yielded the highest P. Unlike the results on the MELB dataset, all variants of DS-ResUNet outperformed their vanilla ResUNet baseline in P, IoU and F1 without a significant increase in network parameters from DS-ResUNet-S. The smallest variant DS-ResUNet-S outperforms ResUnet in three evaluation measures with approx. 0.8\% increase in network parameter. Similarly, the largest variant DS-ResUNet-A outperformed ResUnet with a 13.6\% rise in network parameters. Among the DS-UNet3+ variants, DS-UNet3+(L) yielded the highest IoU and F1 but without a significant increase from the vanilla U-Net3+. All proposed networks outperform the majority of the SOTA networks. In particular, the high scorers DS-ResUNet-A and DS-UNet3+(L) outperformed all SOTA networks except the DeepLabv3+ in this dataset. DeepLabv3+ is the high scorer in IoU and F1 for this dataset. 

\begin{table}[!hbt]
\centering
\caption{DS-UNet, DS-ResUNet, and DS-UNet3+ on 1.2m subset of MELB dataset. The highest scores are highlighted in bold in each group of experiments. The top 3 scores are ranked in order with a postscript for each evaluation metric.}
\label{tab:res:melb12}
\centering
\small
\begin{tabular}{lcccc}
\hline
Networks & P & R & IoU & F1 \\ \hline
U-Net & 0.901 & \textbf{0.929} & \textbf{0.842}$^2$ & \textbf{0.912}$^2$ \\
DS-UNet-L & 0.932$^2$ & 0.893 & 0.839 & 0.909 \\
DS-UNet-S & \textbf{0.944}$^1$ & 0.880 & 0.837 & 0.908 \\
DS-UNet-A & 0.925 & 0.903 & 0.841$^3$ & 0.911$^3$ \\ \hline
ResUnet & 0.856 & \textbf{0.933} & 0.806 & 0.889 \\
DS-ResUNet-L & 0.884 & 0.927 & 0.827 & 0.902 \\
DS-ResUNet-S & 0.882 & 0.932 & 0.829 & 0.903 \\
DS-ResUNet-A & \textbf{0.903} & 0.914 & \textbf{0.832} & \textbf{0.905} \\ \hline
U-Net3+ & 0.903 & 0.921 & 0.837 & \textbf{0.909} \\
DS-UNet3+(L) & 0.866 & 0.964$^3$ & \textbf{0.838} & \textbf{0.909} \\
DS-UNet3+(S) & 0.856 & \textbf{0.967}$^2$ & 0.831 & 0.904 \\
DS-UNet3+(A) & \textbf{0.927}$^3$ & 0.893 & 0.835 & 0.907 \\ \hline
U-Net++ & 0.881 & 0.879 & 0.789 & 0.877 \\
TransUnet & 0.848 & 0.940 & 0.801 & 0.886 \\
Swin-Unet & 0.876 & 0.909 & 0.805 & 0.889 \\
Att.Unet & 0.809 & 1.000$^1$ & 0.809 & 0.887 \\
U$^2$Net & 0.904 & 0.901 & 0.823 & 0.899 \\
FCN8s & 0.892 & 0.628 & 0.592 & 0.729 \\
DeepLabv3+ & 0.922 & 0.917 & 0.852$^1$ & 0.917$^1$ \\
SegNet & 0.841 & 0.927 & 0.793 & 0.878 \\ \hline
\end{tabular}
\end{table}

\subsection{Results on high-resolution dataset (Massachusetts)} \label{sec:res:massachusetts}
\begin{figure}[!ht]
  \centering
  \includegraphics[width=\linewidth]{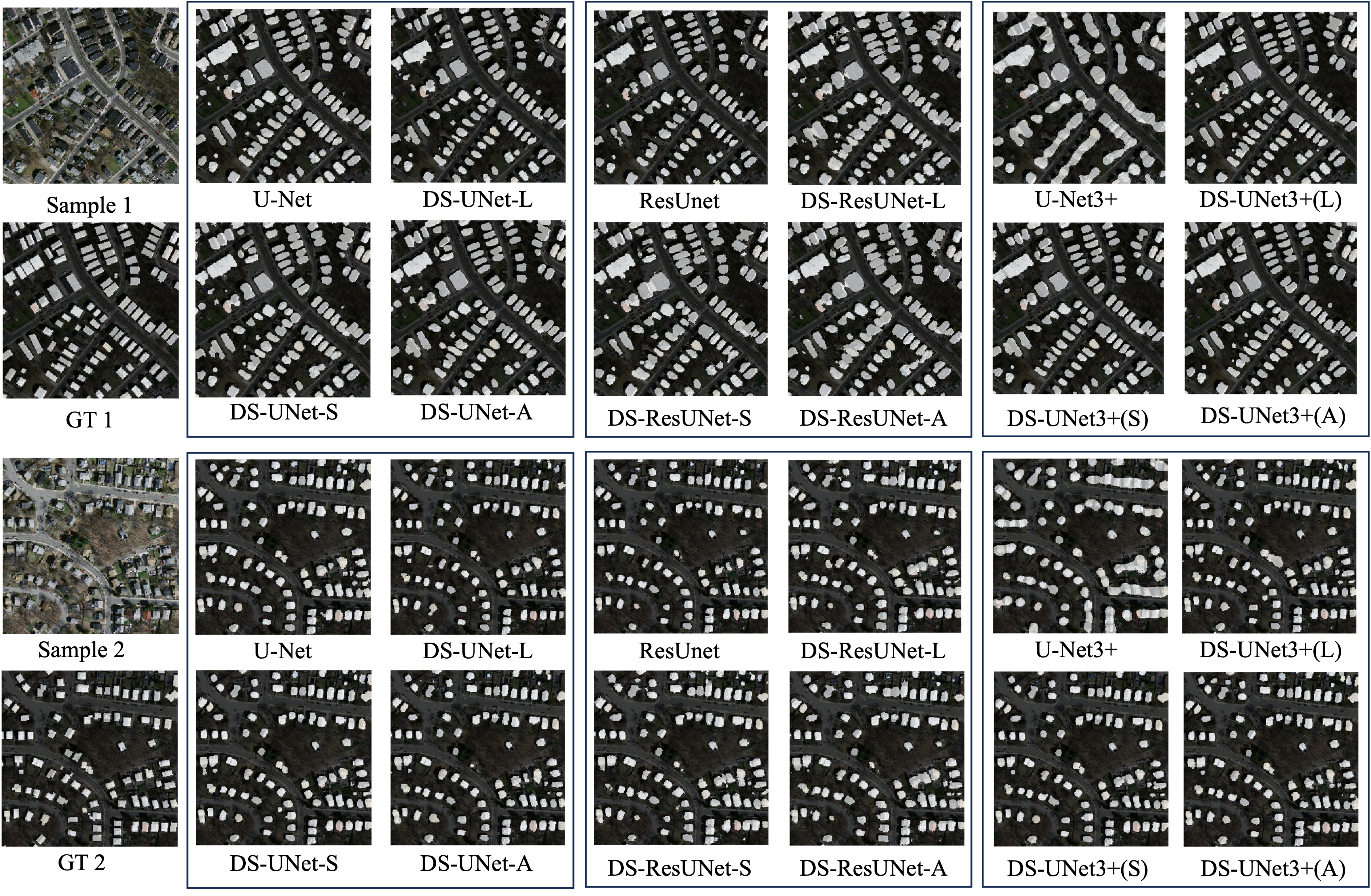}
  \caption{Segmentation output from the proposed networks and their original vanilla baseline networks on the Massachusetts building dataset.}
      \label{fig:MASS-results}
      \end{figure}

Table \ref{tab:res:mass} presents the results on the high-resolution Massachusetts building dataset. Among the DS-UNets, DS-UNet-L and DS-UNet-S outperform U-Net, with the first one being the high scorer among its variants. The variants of DS-ResUNet did not outperform ResUnet in R, IoU, and F1. In DS-UNet3+ variants, DS-UNet3+(L) and DS-UNet3+(A) outperformed U-Net3+ in P, IoU, and F1. The proposed networks outperformed the majority of the SOTA networks. Among all networks, U$^2$Net is the high scorer in IoU and F1 for this dataset. Figure \ref{fig:MASS-results} shows the segmentation results of all proposed and original vanilla networks. 

\begin{table}[!hbt]
\centering
\caption{DS-UNet, DS-ResUNet, and DS-UNet3+ on 1m high-resolution Massachusetts building dataset. The highest scores are highlighted in bold in each group of experiments. The top 3 scores are ranked in order with a postscript for each evaluation metric.}
\label{tab:res:mass}
\centering
\small
\begin{tabular}{lcccc}
\hline
Networks & P & R & IoU & F1 \\ \hline
U-Net & 0.961$^2$ & 0.937 & 0.905 & 0.948 \\
DS-UNet-L & 0.945 & \textbf{0.955} & \textbf{0.907} & \textbf{0.949}$^3$ \\
DS-UNet-S & 0.959$^3$ & 0.941 & 0.906 & 0.949 \\
DS-UNet-A & \textbf{0.962}$^1$ & 0.936 & 0.906 & 0.948 \\ \hline
ResUnet & 0.934 & \textbf{0.967}$^2$ & \textbf{0.907}$^3$ & \textbf{0.949} \\
DS-ResUNet-L & \textbf{0.956} & 0.921 & 0.886 & 0.937 \\
DS-ResUNet-S & 0.955 & 0.919 & 0.883 & 0.935 \\
DS-ResUNet-A & \textbf{0.956} & 0.927 & 0.892 & 0.940 \\ \hline
U-Net3+ & 0.838 & \textbf{1.000}$^1$ & 0.838 & 0.907 \\
DS-UNet3+(L) & 0.959 & 0.937 & 0.904 & 0.947 \\
DS-UNet3+(S) & 0.838 & \textbf{1.000}$^1$ & 0.838 & 0.907 \\
DS-UNet3+(A) & \textbf{0.961}$^2$ & 0.938 & \textbf{0.905} & \textbf{0.948} \\ \hline
U-Net++ & 0.952 & 0.911 & 0.874 & 0.930 \\
TransUnet & 0.951 & 0.953 & 0.910$^2$ & 0.951$^2$ \\
Swin-Unet & 0.927 & 0.913 & 0.856 & 0.918 \\
Att.Unet & 0.959 & 0.936 & 0.903 & 0.947 \\
U$^2$Net & 0.949 & 0.958$^3$ & 0.913$^1$ & 0.953$^1$ \\
FCN8s & 0.934 & 0.912 & 0.861 & 0.922 \\
DeepLabv3+ & 0.959 & 0.945 & 0.910$^2$ & 0.951$^2$ \\
SegNet & 0.926 & 0.947 & 0.884 & 0.935 \\ \hline
\end{tabular}
\end{table}

\subsection{Results on VHR building dataset (WHU)} \label{sec:res:VHR}
\begin{figure}[!ht]
  \centering
  \includegraphics[width=\linewidth]{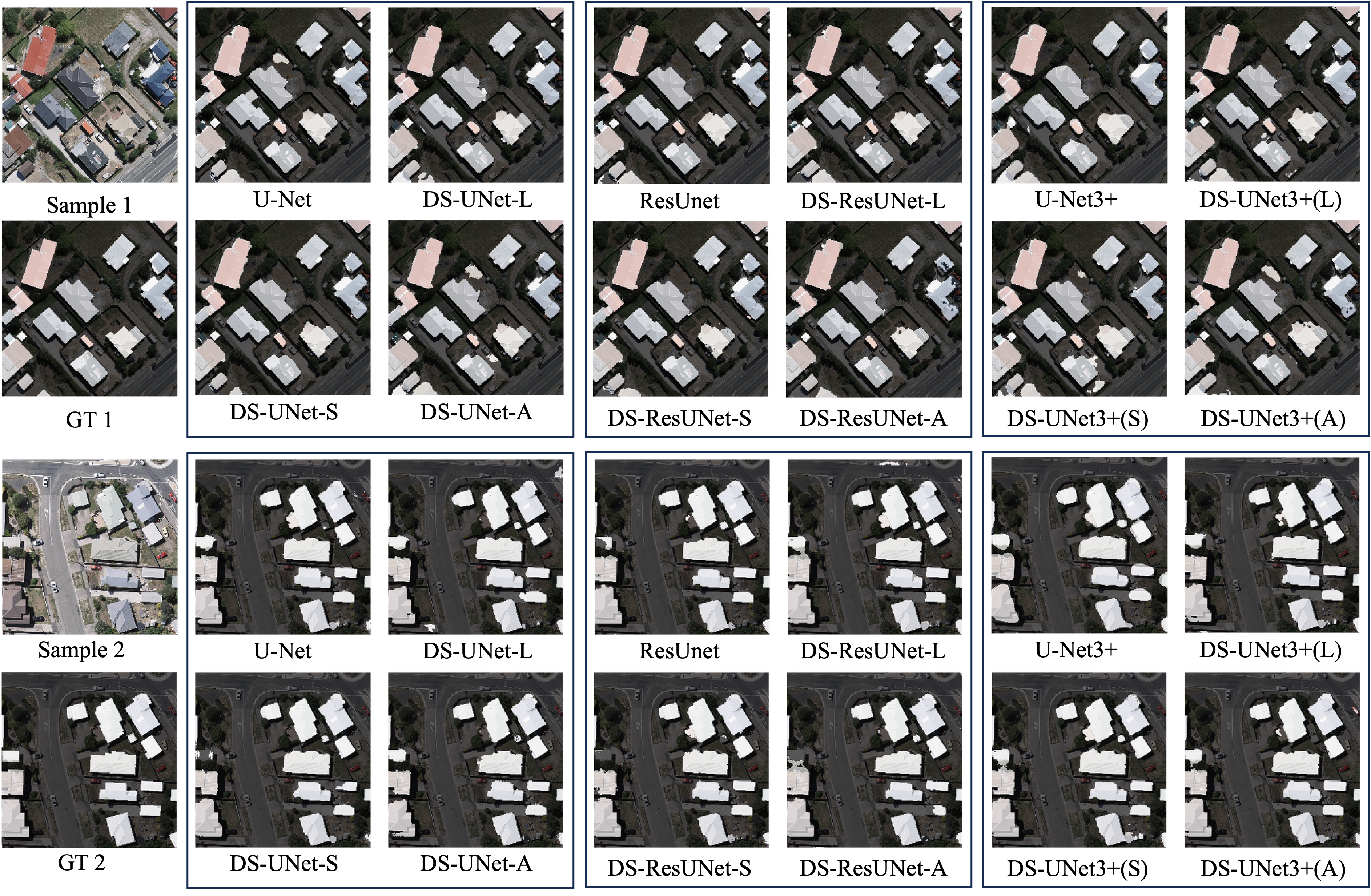}
  \caption{Segmentation output from the proposed networks and their original vanilla baseline networks on the WHU building dataset.}
      \label{fig:WHU-results}
      \end{figure}

Table \ref{tab:res:VHR} presents the performance of three versions of DS-UNet on the VHR benchmark WHU building dataset. Among the DS-UNets, DS-UNet-A produced similar scores as U-Net, without performance gains from the three proposed networks. DS-UNet-L and DS-UNet-A yield marginally higher P and R respectively. For DS-ResUNet, L, S and A respectively produced the highest P, R and IoU but with marginal gains. For DS-UNet3+, all of its three versions outperform the vanilla U-Net3+ with significant performance gains, with the highest IoU and F1 from DS-UNet3+(L). The proposed networks did not yield a higher F1 score than their baselines. Once again, most of the proposed networks yielded higher scores than the majority of SOTA networks. Attention U-Net is the high scorer in IoU and F1 among all networks in this dataset. The sample results are shown in Figure \ref{fig:WHU-results}.

\begin{table}[!hbt]
\centering
\caption{DS-UNet, DS-ResUNet, and DS-UNet3+ on 0.3m VHR WHU building dataset. The highest scores are highlighted in bold in each group of experiments. The top 3 scores are ranked in order with a postscript for each evaluation metric.}
\label{tab:res:VHR}
\small
\centering
\begin{tabular}{lcccc}
\hline
Networks & P & R & IoU & F1 \\ \hline
U-Net & 0.987 & 0.982 & \textbf{0.970} & \textbf{0.983} \\
DS-UNet-L & \textbf{0.990} & 0.969 & 0.960 & 0.978 \\
DS-UNet-S & 0.989 & 0.978 & 0.967 & 0.982 \\
DS-UNet-A & 0.983 & \textbf{0.985} & 0.969 & \textbf{0.983} \\ \hline
ResUnet & 0.991$^2$ & 0.982 & 0.973$^3$ & \textbf{0.986}$^2$ \\
DS-ResUNet-L & \textbf{0.993}$^1$ & 0.978 & 0.972 & 0.985$^3$ \\
DS-ResUNet-S & 0.984 & \textbf{0.987}$^3$ & 0.972 & 0.985$^3$ \\
DS-ResUNet-A & 0.987 & 0.986 & \textbf{0.974}$^2$ & \textbf{0.986}$^2$ \\ \hline
U-Net3+ & 0.889 & \textbf{1.000}$^1$ & 0.889 & 0.934 \\
DS-UNet3+(L) & \textbf{0.989} & 0.983 & \textbf{0.973}$^3$ & \textbf{0.985}$^3$ \\
DS-UNet3+(S) & 0.991$^2$ & 0.975 & 0.968 & 0.982 \\
DS-UNet3+(A) & 0.987 & 0.984 & 0.972 & \textbf{0.985}$^3$ \\ \hline
U-Net++ & 0.989 & 0.980 & 0.971 & 0.984 \\
TransUnet & 0.990 & 0.983 & 0.974$^2$ & 0.986$^2$ \\
Swin-Unet & 0.979 & 0.956 & 0.938 & 0.966 \\
Att.Unet & 0.986 & 0.989$^2$ & 0.975$^1$ & 0.987$^1$ \\
U$^2$Net & 0.989 & 0.971 & 0.962 & 0.979 \\
FCN8s & 0.977 & 0.945 & 0.926 & 0.956 \\
DeepLabv3+ & 0.989 & 0.984 & 0.974$^2$ & 0.986$^2$ \\
SegNet & 0.990$^3$ & 0.968 & 0.960 & 0.978 \\ \hline
\end{tabular}
\end{table}

\subsection{Limitations} \label{sec:res:limit}
A uniform experimental setup is used to compare the proposed networks, the original vanilla networks, and other SOTA networks. The experiments are designed to focus on the performance gain in the segmentation of VHR and high-resolution EO images. No pre-processing and post-processing methods are used for the regularisation of building boundaries. The MELB dataset is prepared end-to-end without manual annotations, with images from an API service of Nearmap and labels from a secondary source. However, due to the UAV-based sources, the dataset is affected by the off-nadir angle of source images. This problem results in a misalignment between images and labels, which rises along with the height of the buildings. The misalignment further rises in the higher spatial resolution subset if measured in pixels. This means that in a 256x256 tile, there is more misalignment between an object (building) and its label in 0.3m spatial resolution when compared to a 1.2m resolution image. Therefore, training only on a lower-resolution dataset might not result in a higher performance score, but it definitely assists in the generalisation of objects if bundled together with a higher-resolution dataset. The majority of the publicly available building footprint datasets cover residential buildings with no complex high-rises and skyscrapers. The end-to-end prepared multi-resolution MELB dataset could therefore be useful for the precise and automated extraction of urban buildings.

\section{Conclusion} \label{sec:conclusion}
This paper dissects and re-thinks the network configuration of three widely used SOTA EDNs for building footprint extraction. The particular focus is on re-designing the skip connections of U-Net, ResUnet, and U-Net3+ that transport the learned vectors from their CNN encoder to their decoder. Three dual skip connection mechanisms $-$ DSCM, DRSCM, and DFSCM $-$ are designed for U-Net, ResUnet, and U-Net3+ respectively to compensate for their limitations and to achieve performance gain. Moreover, these mechanisms provide an extensive experimental study to find the selective skip connections that can be densified to achieve higher accuracy measures. This further provides a more accurate trade-off between the use of context and precise localisation for semantic segmentation of VHR and high-resolution EO images. Further, a dual full-scale skip connection mechanism (DFSCM) is proposed for the U-Net3+. DFSCM incorporates the increased low-level context information with high-level semantics from the feature maps at different scales using a multi-scale feature aggregation technique. Unlike U-Net3+, the aggregation proposed aggregates the multi-scale feature maps with increased weights for the smaller-scale feature maps, where the context information is still intact.

The three proposed mechanisms double the feature maps in the encoder before passing them to the decoder and they can be plugged into specific scale layers of the networks. With this advantage, the dual skip connection mechanism is tested in different scale layers of three networks, resulting in a total of nine network configurations. Furthermore, an ablation study is provided to confirm the optimal scale layer for performance gain. The following conclusions can be made from our experiments on VHR and high-resolution building footprint datasets:

\begin{enumerate}
    \setcounter{enumi}{0}
    \item Enriching the plain skip connections of U-Net with increased large-scale features from DSCM provides performance gain in extracting buildings from the multi-resolution MELB dataset and the high-resolution Massachusetts Building dataset. 
    \item The feature maps of the third encoder layer can be deepened for denser third skip connection to achieve performance gain without needing to densify all the skip connections in U-Net.
    \item The residual blocks of ResUnet can be deepened with DRSCM for performance gain in some datasets like VHR WHU Building dataset and 1.2m subset of the MELB dataset.
    \item DS-UNet3+ variants yield a higher F1 and IoU with increased feature maps using DFSCM and DSFAM aggregation in the majority of datasets. Therefore, assigning different weights to the differently sized feature maps of varying levels of discrimination allows performance gain in U-Net3+.
    \item The proposed DS-UNets and DS-UNet3+ networks outperform their vanilla baselines and the majority of SOTA networks on the proposed MELB dataset and its subset that consists of complex urban buildings with buildings ranging from residential low-rise and mid-rise buildings to high-rise and skyscrapers. The majority of the proposed networks also outperform Transformer-based networks $-$ TransUnet and SwinUnet. In the MELB dataset, DS-UNet3+(S) outperforms TransUnet with almost 19 times lower network parameters.
\end{enumerate}

The studies from this paper highlight several gaps in the literature that could be addressed in future works. The gaps are: (i) lack of studies to point out the optimal scale layers that need skip connections, (ii) the majority of EO-based high-resolution building datasets cover residential areas resulting in a lack of studies in complex urban settings that include high-rise and skyscrapers, (iii) lack of conclusions that point out the optimal resolution of EO images for urban feature extraction, and (iv) the misalignment between roof labels and UAV-based aerial images due to inaccurate ortho-rectification methods, resulting in inaccurate urban building footprint extraction. The contributions in this paper touch on some of these gaps, however, our future works will be dedicated to addressing and minimising the other gaps.

\backmatter


\bmhead{Acknowledgments}
The first author (B.N.) is supported by the University of Melbourne for his Ph.D. research and has been awarded by Melbourne Research Scholarship. This research was undertaken using the LIEF HPC-GPGPU Facility hosted at the University of Melbourne, which was established with the assistance of LIEF Grant LE170100200. The authors would like to thank Nearmap for providing the API service to collect the image data for the experiments.

\bmhead{Data Availability}
Data deposited in a repository: \\ \href{https://github.com/bipulneupane/DualskipUnets/}{https://github.com/bipulneupane/DualskipUnets/}.

\section*{Declarations}
\begin{itemize}
\item Funding: No funding was received for this study.
\item Conflict of interest: The authors declare no conflict of interest.
\item Code availability: The codes will be available in a GitHub link after peer review.
\end{itemize}







\bibliography{sn-bibliography}

\end{document}